\documentclass{article}
\usepackage{ijcai25}

\usepackage{times}
\usepackage{soul}
\usepackage{url}
\usepackage[hidelinks]{hyperref}
\usepackage[utf8]{inputenc}
\usepackage[small]{caption}
\usepackage{graphicx}
\usepackage{caption} 
\usepackage{amsmath}
\usepackage{amsthm}
\usepackage{booktabs}
\usepackage{algorithm}
\usepackage{algorithmic}
\usepackage{bbding}
\usepackage{booktabs}
\usepackage{hyperref}
\usepackage{amssymb}

\title{FPGA: Flexible Portrait Generation Approach}

\author{
Zhaoli Deng,
Fanyi Wang$^*$,
Junkang Zhang,
Fan Chen,
Meng Zhang,
Wendong Zhang,
Wen Liu$^*$,
Zhenpeng Mi$^*$\\
\affiliations
 Honor Device Co., Ltd\\
\emails
wangfanyi@honor.com
}

\begin{document}

\twocolumn[{%
\renewcommand\twocolumn[1][]{#1}%
\maketitle
\begin{center}
    \captionsetup{type=figure}
    \includegraphics[width=1.0\linewidth]{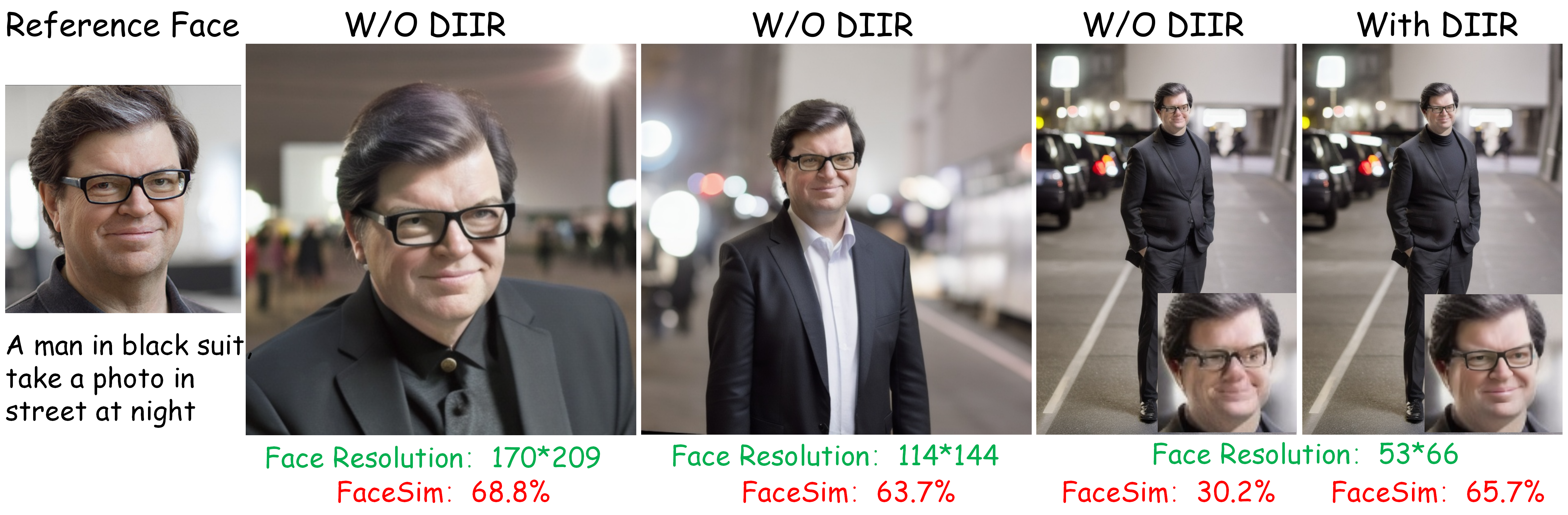}
    \captionof{figure}{Generation results of \textbf{FPGA} with different resolution faces, and face restoration capabilities of our \textbf{DIIR}.}
    \label{fig:faceResolution}
\end{center}%
}]

\begin{abstract}
Portrait Fidelity Generation is a prominent research area in generative models. 
Current methods face challenges in generating full-body images with low-resolution faces, especially in multi-ID photo phenomenon.
To tackle these issues, we propose a comprehensive system called FPGA and construct a million-level multi-modal dataset IDZoom for training. 
FPGA consists of Multi-Mode Fusion training strategy (MMF) and DDIM Inversion based ID Restoration inference framework (DIIR).  
The MMF aims to activate the specified ID in the specified facial region. 
The DIIR aims to address the issue of face artifacts while keeping the background. 
Furthermore, DIIR is plug-and-play and can be applied to any diffusion-based portrait generation method to enhance their performance. DIIR is also capable of performing face-swapping tasks and is applicable to stylized faces as well. 
To validate the effectiveness of FPGA, we conducted extensive comparative and ablation experiments. 
The experimental results demonstrate that FPGA has significant advantages in both subjective and objective metrics, and achieves controllable generation in multi-ID scenarios. In addition, we accelerate the inference speed to within 2.5 seconds on a single L20 graphics card mainly based on our well designed reparameterization method, RepControlNet.
\end{abstract}

\section{Introduction}
Portrait image generation based on the text-to-image diffusion model~\cite{rombach2022high} has significant commercial value and is a hot topic for research. 
Early works require multiple images of a same ID from different angles. 
By fine-tuning the diffusion model~\cite{dreambooth} or regularization modules~\cite{hu2021lora}, these methods ultimately obtain a customized model for a specific ID. 
Although these methods can successfully insert the ID information into the customized model, the requirements for multiple images and model training still limit the application of these methods.
The introduction of IP-Adapter~\cite{ye2023ip} partially addressed these issues by extracting image embeddings from a single reference image and incorporating them into the denoising process through weighted cross-attention~\cite{vaswani2017attention}, which reduced the demand for multiple images.
%
To further improve the ID fidelity in portrait image synthesis, the authors of IP-Adapter replace the general image vision encoder with a specific image encoder pre-trained on the face classification task, and propose the IP-Adapter-Face. Thereafter, many excellent works emerged to generate highly faithful ID preserved images from a single portrait image, such as PhotoMake~\cite{li2023photomaker}, InstantID~\cite{wang2024instantid}, ConsistentID~\cite{huang2024consistentid}, PuLID~\cite{guo2024pulid} and InstantFamily~\cite{InstantFamily}.
%
Although these works have achieved impressive results for portrait image generation, generating multi-ID full-body images with specified positions remains a challenging problem.
On one hand, methods like FasterComposer~\cite{xiao2023fastcomposer} can generate multi-ID photos but fail to specify the positions. Existing methods~\cite{wang2024instantid,li2023photomaker} achieve the goal of position-specific generation by conducting Mask Guided Multi-ID Cross-Attention during the inference stage as shown in Fig.~\ref{fig:MFF}, but they may well fail to activate specific ID at specific face region as shown in Fig.~\ref{fig:prove_CFT}.
In addition, current portrait fidelity methods still fail to generate fine-grained facial details in scenarios where the face resolution is small, as shown in Fig.~\ref{fig:faceResolution}. The obvious artifacts significantly limit the practical applications of these methods.
To address these issues, we propose a novel ID fidelity generation approach, named FPGA, which consists of the Multi-Mode Fusion Training Strategy (MMF) and the DDIM Inversion-based ID Restoration inference framework (DIIR). As its name implies, its inference speed is fast. Our contributions are as follows.
\begin{itemize}
\item For multi-ID position-specific generation, we propose the Multi-Mode Fusion training strategy (MMF), and introduce Clone Face Tuning training strategy and Mask Guided Multi-ID Cross Attention. 
\item To address the issue of face artifacts, we propose the DDIM Inversion based ID Restoration inference framework (DIIR), which ensures facial detail restoration while maintaining background consistency. In particular, DIIR is plug-and-play and can be applied to any diffusion-based portrait generation methods which has been proved in Table.\ref{tab:Quantitative_fullbody}. DIIR is also capable of performing face-swapping tasks and is even applicable to stylized faces, results are given in supplementary materials. 
\item In addition, we have compressed the inference latency into 2.5 seconds on a single L20 graphics card mainly by designing a ControlNet reparameterization method, RepControlNet.
\item We spent a lot of effort on constructing the IDZoom dataset, containing millions of samples with six modalities: caption, body, skeleton, face, landmark, and mask of the face region.
\end{itemize}

\section{Related Work}
The following sections primarily introduce works related to Single-ID and Multi-ID personalization.

\subsection{Single-ID Personalization}
Initially, fine-tuning is required to  models, such as LoRA~\cite{hu2021lora} and Dreambooth~\cite{dreambooth}, to generate personalized characters, with limited training concepts.
Therefore, methods like ~\cite{xiao2023fastcomposer,Taming,instance0} use facial models to extract facial features and introduced them into diffusion models through cross-attention, and retrain the SD UNet on portrait data. Achieve the ability of generating target character based on a single portrait image during inference. 
Subsequently, IP-Adapter~\cite{ye2023ip} decouples text conditions and image conditions, only trains the adapter module while freezing the SD model. Thus, IP-Adapter can adapt to different fundation SD models and ControlNet, achieving richer and more controllable personalized generation.
InstantID and Face-Adapter\cite{wang2024instantid,han2024face} focus on controlling facial shapes and enhance facial similarity by introducing more spatial information. 
The second category aims to restore facial details and improve the model's identity retention capability by introducing stronger encoders~\cite{zhang2024flashface}, more reference facial images~\cite{li2023photomaker}, and finer-grained facial descriptions~\cite{huang2024consistentid}. 
Other works concentrate on balancing personalized editing and ID retention, such as ~\cite{guo2024pulid}, which introduces more training losses to decouple text and identity features, and ~\cite{wu2024infinite}, which optimizes inference schemes to achieve more stylized generation. 
However, existing works focus on generating portrait close-ups, neglecting the issue of facial artifacts when generating full-body images.
\subsection{Multi-ID Personalization}
Unlike single-ID generation, multi-ID generation requires generating different characters within the same image and specifying their respective positions. Many single-ID generation methods use segmentation regions during the inference phase to generate multi-ID images. However, this approach can lead to identity confusion or loss. FastComposer~\cite{xiao2023fastcomposer} incorporates location information into the training process, but it cannot specify the positions of the generated characters. On the other hand, InstantFamily~\cite{InstantFamily} introduces spatial information to guide the generation, enabling the specification of character positions. However, multi-ID datasets are difficult to collect, and the small proportion of facial regions makes it challenging to train the identity preserving capability.
\subsection{Accelerating Diffusion Models}

In the aspect of accelerating diffusion models, SnapFusion \cite{li2024snapfusion} focuses on model compression by reducing the amount of model parameters and deploying distillation techniques to enhance inference speed. Step distillation accelerates the inference process by compressing the number of inference steps. The core idea of step distillation is to train a teacher model to generate intermediate step targets, and then use these targets to train a student model, thereby reducing the number of steps required for inference. For example, Consistency Models \cite{song2023consistency} introduce consistency constraints during training to reduce inference steps while maintaining generation quality. Progressive Distillation \cite{salimans2022progressive} adopts a progressive distillation approach, optimizing inference steps incrementally to achieve efficient inference acceleration. However, these methods often come at the cost of sacrificing some generation quality.


Reparameterization techniques optimize model architecture during both training and inference stages. Complex model structures are used during training to enhance the model's expressive power, while these complex structures are reparameterized into simpler forms during inference to reduce computational complexity. RepVGG \cite{ding2021repvgg} and RepViT \cite{Wang_2024_CVPR} are typical applications of reparameterization techniques. RepVGG reparameterizes complex convolutional layers used during training into simple convolutional layers for inference, thereby reducing computational complexity. RepViT applies this technique to Vision Transformers, significantly improving inference speed while maintaining model performance. However, these methods do not specifically address the issue of controllable generation.


Although these methods have made significant contributions to accelerating diffusion models, they typically focus only on the model itself and do not consider the additional complexity introduced by controllable generation. Inspired by RepVGG and ControlNet, we propose a modal reparameterization method, RepControlNet, to address the issue of increased computational overhead in ControlNet. During the training process, RepControlNet freezes the original diffusion model and duplicates its convolutional and linear layers to learn the input modal control information. During inference, the duplicated convolutional and linear layers are reparameterized with the original diffusion model, achieving conditional controllable generation without adding extra computational load beyond the diffusion model itself.

\begin{figure*}[t]
\begin{center}
\includegraphics[width=1.0\textwidth]{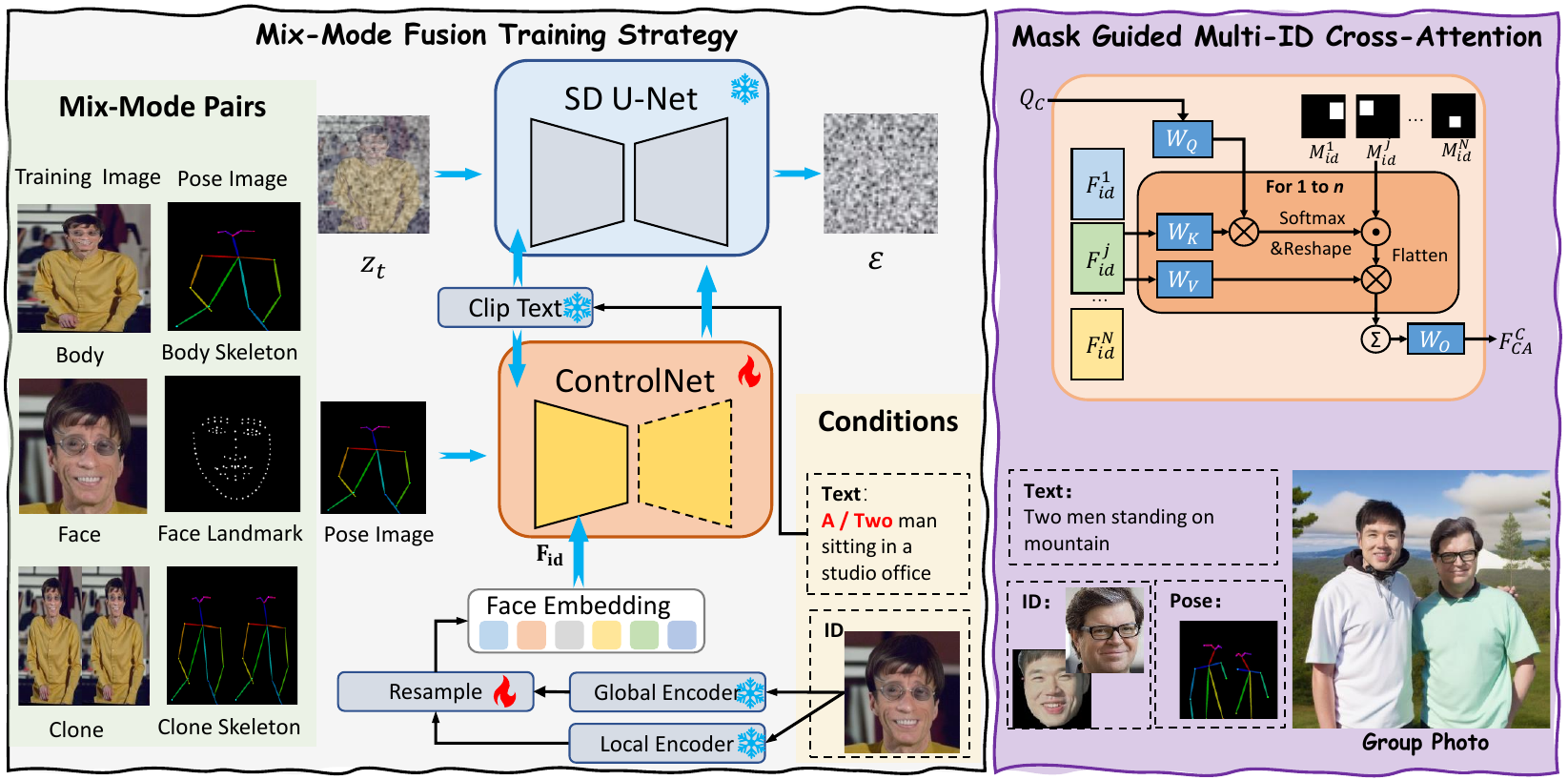}
\end{center}
   \caption{Flowchart of Multi-Mode Fusion training strategy (MMF), the clone operation in sec.~\ref{sec:3.1.3} enables the Mask Guided Multi-ID Cross Attention in inference process to achieve the ability of specifying the location of specified face. }
   \label{fig:MFF}
\end{figure*}

\begin{figure*}[t]
\begin{center}
\includegraphics[width=1.0\textwidth]{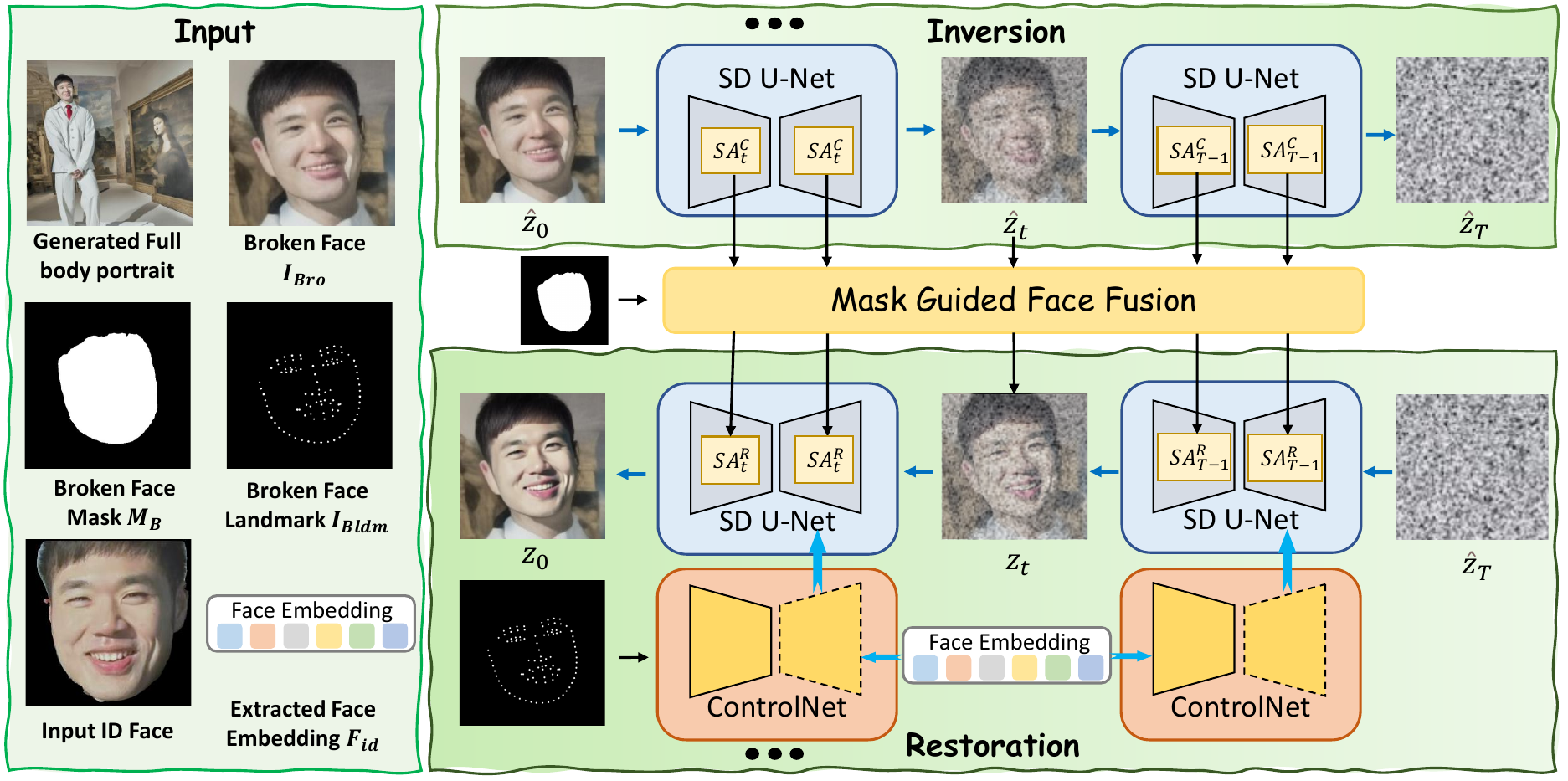}
\end{center}
   \caption{Flowchart of DDIM Inversion based ID Restoration inference framework (DIIR). DIIR achieves crash face repairment.}
   \label{fig:DIIR}
\end{figure*}

\section{Methods}

Our \textbf{FPGA} aims to achieve the ability of generating multi-ID image with clear faces even if the face resolution is extremely small. FPGA consists of two main procedures, training and inference separately, we will introduce the Multi-Mode Fusion Training Strategy in sec.~\ref{sec:3.1}, DDIM Inversion based ID Restoration inference framework (DIIR) in sec.~\ref{sec:3.2}, and RepControlNet in sec.~\ref{sec:RepControlNet}. 

\subsection{Multi-Mode Fusion Training Strategy}
\label{sec:3.1}
The Multi-Mode Fusion training strategy (MMF) utilizes multi-modal data for training, enabling the model to synthesize realistic human portraits conditioned on given prompt, condition image (skeleton, landmark) and identity image.

\subsubsection{Dataset IDZoom}
We collected open source portrait datasets CeleB-A~\cite{liu2015faceattributes}, FFHQ~\cite{karras2019style}, LAION-Face~\cite{zheng2022general}, and images from Internet, about 25 million portrait images in total. Then, those with low resolution and clarity, multi-faces, repeatability are filtered out utilizing the automated toolchain constructed by ourselves. For detail, we use the Openpose~\cite{8765346} to detect the body skeleton, apply the insightface~\cite{deng2020subcenter,deng2018arcface} to detect the face, and PIPNet~\cite{JLS21} to detect the face landmarks. We also conducted manual verification on the data after the initial screening. Considering the efficiency, we selectively performed manual cleaning based only on the skeleton and landmark. The manual cleaning process took approximately 10 person-days in total. Finally, we obtain a high-quality dataset named IDZoom, which contains around 1 million data pairs with six modalities: caption, human image, skeleton, face image, face landmark, and face region mask. 


\subsubsection{Model Architecture}
The architecture of our FPGA is simple yet effect, the flowchart is illustrated in Fig.~\ref{fig:MFF}, and the key point is the following training technique. The fundation T2I diffusion model is freezed, so that the diffusion model can be flexibly changed for updating and more convenient adaptation for the open-source community. We initialize ControlNet with the parameters of the pre-trained stable diffusion UNet. ControlNet encodes condition image (skeleton, landmark) by adding residuals to the corresponding layers of the diffusion model, so as to add spatial control to the fundation T2I diffusion model. For identity preservation, we modify the cross-attention module of ControlNet. For detail, we utilize global Clip Vision ~\cite{DBLP:journals/corr/abs-2103-00020} and ArcFace encoders ~\cite{deng2018arcface} to extract fine-grained face features. Then resample these features into face embedding $F_{id}$ with the dimension of 4$\times$768. $F_{id}$ is injected into ControlNet by cross-attention, the specific operation is shown in Eq.~\ref{crossattn_id_func},
\begin{equation}
\label{crossattn_id_func}
\begin{aligned}
CA^C = CA^C\left(Q_C, F_{text} \right) + CA^C\left(Q_C, F_{id}\right),\\
\end{aligned}
\end{equation}
where $CA^C$ is the cross-attention operation. $Q_C$ is the attention query map of ControlNet which contains condition (skeleton, landmark) information, $F_{id}$ contains fine-grained face features, and $F_{text}$ is the text embedding.

\label{sec:3.1.3}
\subsubsection{Training Details}
Clone Face Tuning is a specifically designed training technique for Mask Guided Multi-ID Cross Attention, aims to specify the locations of specific faces when generating multi-ID photo. Its mechanism is to enable each ID embedding to be activated in each specified spatial region of cross-attention feature $F_{CA}^C$, so that our FPGA can effectively generate the specified ID at specific region during the inference stage. As illustrated in Fig.~\ref{fig:prove_CFT}, the effectiveness has been proved in experiments. 
For detail, during training process, we duplicate the identity image and splice it to the original image, the same operation is applied to the corresponding pose image, while keeping the identity embedding constant. This operation aims to obtain similar face attention maps in both face regions. To constrain this training process, we further propose the clone face attention loss $L_{cfa}$ for supervising. Firstly, we extract region masks, $M_{1}$ and $M_{2}$, which are of the same shape. Subsequently, we crop the facial features from attention maps in latent space with an adaptive resolution (greater than 32$\times$32). Then we multiply the mask with corresponding feature maps. The detail operation is shown in Eq.~\ref{CFAloss},
\begin{equation}
\begin{aligned}
\label{CFAloss}
   L_{cfa} = \left\|(Crop(F_{CA}^C,M_{1})-Crop(F_{CA}^C,M_{2}))\odot G \right\|_{2}^{2}.
\end{aligned}
\end{equation}
$Crop$ presents crop operation, $F_{CA}^C$ is the cross-attention feature map, $G$ is a Gaussian kernel with the same size as $M_{1}$. To mitigate the potential degradation in image quality due to this duplication, we apply Clone Face Tuning once per 10 iterations. The final training objective function $L$ is as follows:
\begin{equation}
\label{condition_func}
   L_{cldm}=\mathbb{E}_{\boldsymbol{Z}_{0}, \boldsymbol{\epsilon}, t}\left[\left\|\boldsymbol{\epsilon}-\boldsymbol{\epsilon}_{\theta}\left(\boldsymbol{Z}_{t},t, \mathcal{C}, F_{id},I \right) \right\|_{2}^{2}\right],
\end{equation}
\begin{equation}
\label{totalloss}
L =
\left\{
\begin{array}{ll}
&L_{cldm}, \\
&L_{cldm} + \lambda *L_{cfa}\text{, every 10 iterations,}\\
\end{array}\right.
\end{equation}
$\epsilon_{\theta}$ represents our target model, $\mathcal{C}$ is the text prompt, $I$ denotes condition image, $\lambda$ sets to 0.2 during training.

\subsubsection{Mask Guided Multi-ID Cross Attention}
When generating multi-ID photos, the input condition image includes multiple body skeletons, from which facial region masks are extracted automatically. Attributed to the above training technique, different ID embeddings are explicitly activated in specific facial regions. The detail process is illustrated in top right corner of Fig.~\ref{fig:MFF}. As illustrated in the last low of Fig.~\ref{fig:multiID}, in the multi-ID inference process, FPGA can achieve the activation of ID embeddings $F_{id}$ in the corresponding face regions correctly, more results can refer to supplementary materials.

\begin{algorithm}[tb]
\caption{DDIM Inversion based ID Restoration}
\label{alg:diir}
\textbf{Input}: Broken face $I_{Bro}$, broken face parsing mask $M_{B}$, broken face landmark $I_{Bldm}$, face identity embedding $F_{id}$.\\
\textbf{Parameter}: $\hat{Z}_{0}$ is the latent code of $I_{Bro}$, $\hat{Z}$ = $\{\hat{Z}_{0},...,\hat{Z}_{T}\}$ and ${Z}$=[${Z}_{0},...,{Z}_{T}$] are the inversion and forward latent codes, $SA^{C}$=$\{SA^{C}_{1},...,SA^{C}_{T}\}$ and $SA^{R}$=$\{SA^{R}_{1},...,SA^{R}_{T}\}$ are cashed and restored self-attention features separately, $\varepsilon_{\theta}$ means DDIM Sampling.\\
\textbf{Output}: Restored face image $I_{R}$.\\
\begin{algorithmic}[1] 
\STATE $\hat{Z}_{0} \gets VAE_{Encoder}(I_{Bro})$ \\
\STATE $\hat{Z}, SA^{C} \gets DDIM Inversion(\hat{Z}_{0})$
\STATE ${Z}_{T}$=$\hat{Z}_{T}$ \\
\STATE $SA^{R}_{T} \gets \varepsilon_{\theta}\left(Z_{T}, T, \mathcal{C}, F_{id}, I_{ldm}\right)$
\FOR{$t = T$ to $1$ }
    \STATE $SA_{t}^{R} \gets \text{AdaIN}(SA^{R}_{t}*M_{B}, SA^{C}_{t}*(1-M_{B}))$
    \STATE $Z_{t-1} \gets \varepsilon_{\theta}\left(Z_{t}, t, \mathcal{C}, F_{id}, I_{ldm}, SA_{t}^{R} \right)$
    \STATE $Z_{t-1} \gets \text{AdaIN}(Z_{t-1}*M_{B}, \hat{Z}_{t-1}*(1-M_{B}))$
\ENDFOR
\STATE \textbf{} $I_{R} \gets VAE_{Decoder}(Z_{0}) $\\
\STATE \textbf{return} $I_{R}$
\end{algorithmic}
\end{algorithm}

\subsection{DDIM Inversion based ID Restoration Inference Framework}
\label{sec:3.2}
Latent diffusion model often produces unpredictable artifacts for portrait generation due to the low resolution of the generated face, which is usually encountered in full-body or group photo scenarios. In this section, we introduce our DIIR as shown in Fig.~\ref{fig:DIIR}, which is specifically designed to restore broken faces while maintaining contextual background consistency, accurate face pose, and harmonious color.

\subsubsection{High-level Feature Preservation} 
As shown in Fig.~\ref{fig:DIIR}, we use landmarks extracted from broken face as condition image for ControlNet to maintain the high-level features such as face pose and expression. Experimentally, existing methods~\cite{JLS21} can effectively extract face landmarks, even the image contain artifacts. 

\subsubsection{Low-level Feature Preservation} 
\label{sec:3.2.2}
Recent works ~\cite{gu2024swapanything,gu2024photoswap,Chung_2024_CVPR} have demonstrated that the low-level feature such as color and background information, are presented in the self-attention features of diffusion UNet. Hence, we extract this information utilizing DDIM inversion ~\cite{DBLP:journals/corr/abs-2010-02502}. The detail DIIR operation is presented by Alg.\ref{alg:diir}, we encode broken face into the latent space to obtain $\hat{Z}_{0}$, and then add noise step by step through DDIM inversion to obtain $\hat{Z}_{T}$. During the inversion process, we cache the self-attention map, denoted as $SA^{C}$=$[SA^{C}_{1},SA^{C}_{2},...,SA^{C}_{T}]$, and inversion latent code $\hat{Z}$=[$\hat{Z}_{0},...,\hat{Z}_{T}$]. The subscript in the notation represents different time steps. These cached features will be introduced into DDIM forward sampling process by Mask Guided Face Fusion, which is presented by line 7 and 9 in Alg.\ref{alg:diir}. $SA^{R}$=$[SA^{R}_{1},SA^{R}_{2},...,SA^{R}_{T}]$ represents the restoration self-attention map during DDIM forward sampling process, and its latent code is ${Z}$=[${Z}_{0},...,{Z}_{T}$]. We employ AdaIN~\cite{huang2017arbitrary} to modulate the $SA^{R}_{t}$ and $SA^{C}_{t}$ features with spatial constraints $M_{B}$. We extract $M_{B}$ using ~\cite{yu2018bisenet} and smooth its boundary with Gaussian kernel. Then, we replace the self-attention map during DDIM forward sampling process $SA^{R}_{t}$ with the feature map fused by AdaIN, to obtain the latent code ${Z}_{t-1}$. After that, we apply MGFF to ${Z}_{t-1}$ and cached $\hat{Z}_{t-1}$ to update the ${Z}_{t-1}$. Finally, we obtain the denoised latent code ${Z}_{0}$, which is decoded by $VAE_{Decoder}$ to obtain the restoration result. For multi-ID restoration, replace $M_{B}$ with $M_{id}^i$, which basically does not affect the overall latency.

\label{sec:RepControlNet}
\subsection{RepControlNet}
\begin{figure}[t]
\begin{center}
\includegraphics[width=0.5\textwidth]{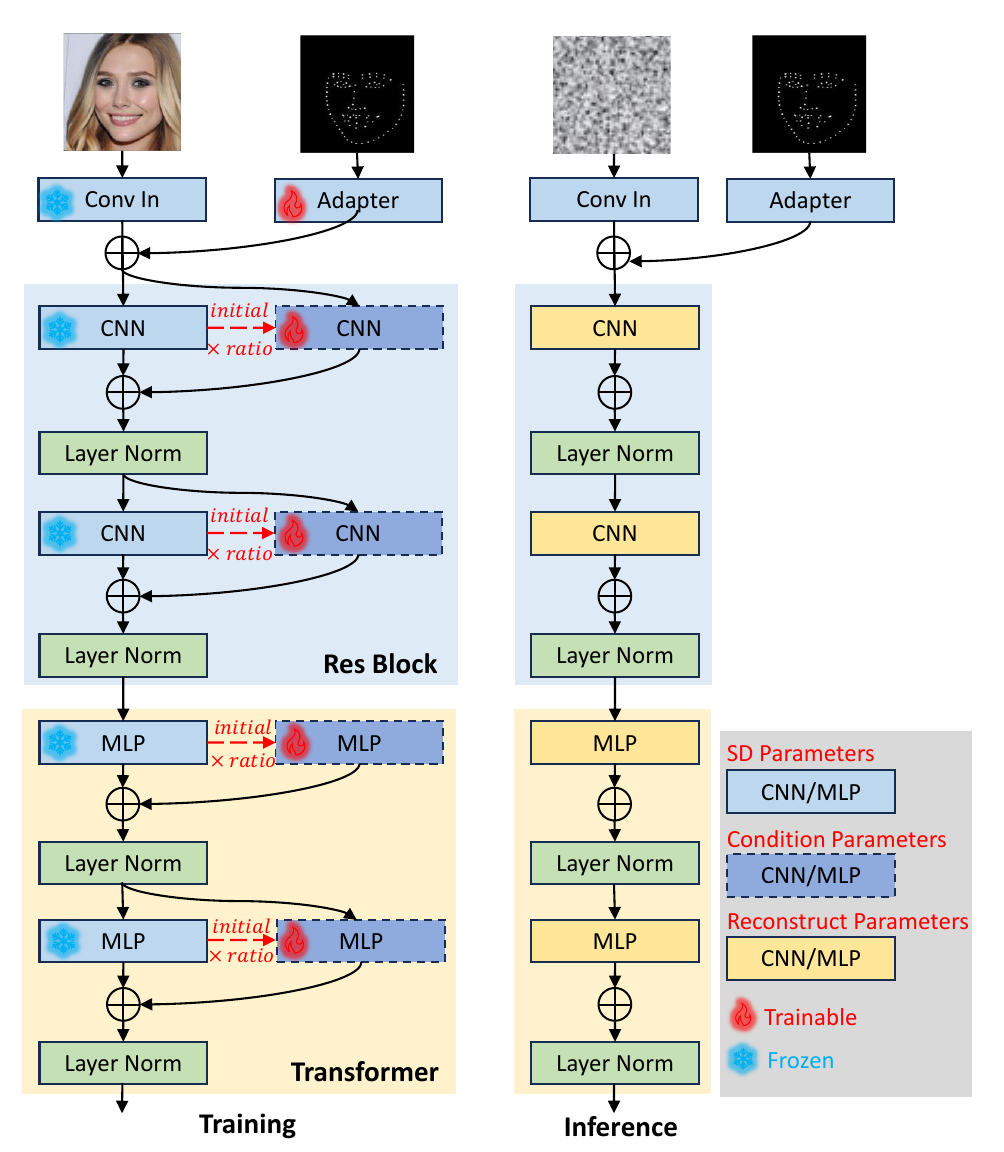}
\end{center}
\caption{Flowchart of training and reparameterization process of RepControNet.}
\label{fig:flowchart}
\end{figure}






We propose a ControlNet reparameterization approach for inference acceleration, called RepControlNet. Similar to ControlNet, the original blocks in diffusion model are frozen in RepControlNet, while a trainable copy is applied to learn conditions. Unlike ControlNet, the trainable copy is only applied to convolution and linear layers in RepControlNet for the convenient reparameterization in inference process.
Specifically, in RepControlNet, all convolution and linear layers are frozen in pretrained diffusion model in training process, then the copy of these layers is trained as Fig.~\ref{fig:flowchart}.
In RepControlNet, the feature embedding is simultaneously input into original layers and copy layers in each convolution and linear layers, then adding the two outputs together as the output feature. 
An Adapter layer is applied to encode the input condition in the first layer of UNet. 
The condition information is injected by adding Adapter output tensor and original input tensor (the two tensors have the same dimension).

In this way, we transfer the single branch model to multi-branch model which achieves the equivalent effect as ControlNet. 
For reusing the strong capacity of pretrained diffusion model and reducing the gap between multi-branch model and original model as much as possible
in the early training process, the weights $\Theta_m$ in multi-branch model is initialized as a multiple of original weights $\Theta$ and a ratio $w$
\begin{equation}
\begin{aligned}
\label{reparameterization}
\Theta_m = w\Theta ,
\end{aligned}
\end{equation}
empirically, the ratio $w$ is set to be $0.1$ in our experiments.

\subsubsection{ReparameteriZation}





The uniqueness of RepControlNet lies in its use of reparameterization technology to achieve efficient conditional controllable generation during the inference process.
In the inference process, RepControlNet simplifies the complex structure during training into an efficient inference structure through reparameterization. 
Specifically, the copy of convolution and linear layers are reparameterized with the weights of the original diffusion model. 
By linearly combining the weights learned during training process with the original weights, conditional controllable generation is achieved without increasing additional computational complexity

Specifically, for each convolution layer and linear layer, RepVGG has demonstrated that by combining the weights of different branches, a multi-branch model can be equivalently transformed into a single-branch model. The weight combination formula in the reparameterization process is 
\begin{equation}
\label{reparameterization}
\begin{aligned}
\Theta' = \alpha \Theta + \beta \Theta_m,
\end{aligned}
\end{equation}
where $\Theta$ is the weight of original diffusion model, $\Theta_m$ is the weight of the copy (modal) model, $\alpha$ and $\beta$ are the ratio hyper-parameters.

During inference, we reparameterize the RepControlNet, whereas the introduction of identity information cannot be reparameterized. However, consistent with text embedding, $F_{id}$ only needs to be extracted once, independent of time steps. Moreover, due to the limited number of tokens in  $F_{id}$, the additional computational overhead for inference is negligible as shown in Table.\ref{table:compute}.


\section{Experiments}

\begin{figure*}[t]
\begin{center}
\includegraphics[width=0.88\textwidth]{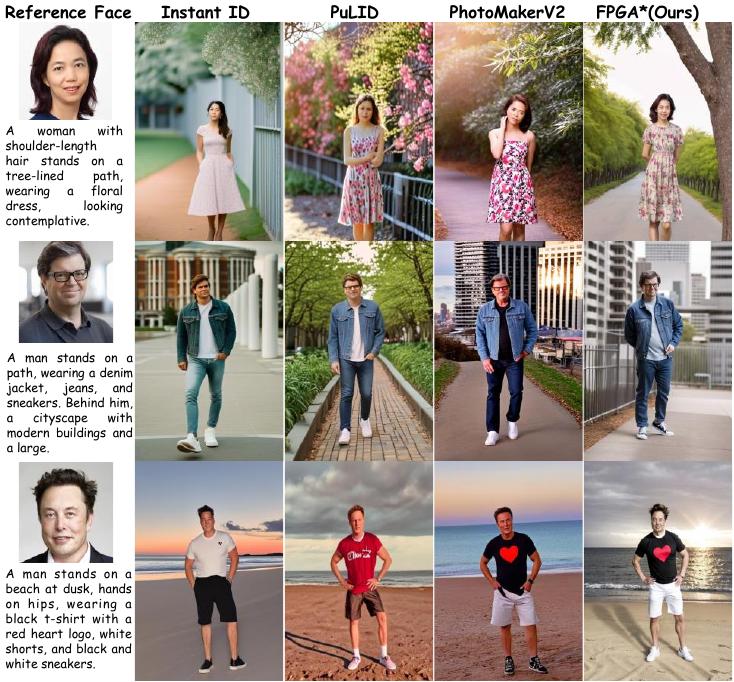}
\end{center}
   \caption{Visualization comparison with SOTA methods on full-body generation ability. FPGA* shows fantastic performance on low resolution face generation, * means using DIIR. More clear results are presented in supplementary materials.}
   \label{fig:full_body_gen}
\end{figure*}

\subsection{Implement Details}
In the experiments, we employ RealisticVisionV4 (SD1.5) as the fundation model. 
The training process was divided into three stages. We first train the ControlNet with face landmark as condition image for 50,000 iterations.
Then, skeleton is utilized as condition image for another 100,000 iterations. The final phase involves training on both condition types for 1,000,000 iterations, with Clone Face Tuning conducted every 10 iterations. 
%
We utilize the Adam optimizer with an initial learning rate of 1e-4, batch size of 128, 8 NVIDIA A800 GPUs, the full training process takes about 2 weeks.
We also set a 10\% chance to replace the text embedding with a null-text embedding to enable classifier-free guidance in the inference stage. 
During inference, we employ 30 steps of DDIM sampler, and classifier-free guidance is set to 5. We execute DIIR operation with $T=30$ when the face resolution of the generated image is less than $100\times100$ pixels.
Additionally, We have greatly accelerated the inference speed through engineering methods such as tensorRT inference, model quantization, deepcache. Ultimately, the inference time is controlled within 2.5 seconds on a single L20 graphics card.
\begin{table}[!t]
\centering\small
  \setlength{\tabcolsep}{0pt} 
  \begin{tabular*}{\columnwidth}{@{\extracolsep{\fill}}lccccccc@{}}
    \toprule
    \textbf{Method} &Arch. &M-ID & ClipI $\uparrow$ & ClipT $\uparrow$ &  FaceSim $\uparrow$ & DINO $\uparrow$  & FID $\downarrow$ \\
    \midrule
    IP-Adapter  &SD1.5 &\checkmark & 89.4 & 19.9 &41.1 &82.4 & 213.8\\
    InstantID  &SDXL &\checkmark &88.7 &19.5  & 57.7  &  79.7 &220.3\\
    PhotomakeV2  &SDXL &\XSolidBrush &88.6  & 20.9  &  50.5 &   81.4 &216.1 \\
    PuLID &SDXL &\XSolidBrush &86.9 & 19.4  & 47.0  &  82.4 & 229.9\\
    \textbf{FPGA(Ours)} &SD1.5 &\checkmark &\textbf{91.1}  & \textbf{21.1}  & \textbf{66.5} & \textbf{87.3} &\textbf{207.9} \\
    \bottomrule
  \end{tabular*}
 \caption{Single-ID half-body comparison. The best average performance is in bold. ↑ indicates higher metric value and represents better performance and vice versa. M-ID means Multi-ID.}
\label{tab:Quantitative_halfbody}
\end{table}

\begin{table}[t]
\centering\small
  \setlength{\tabcolsep}{0pt} 
  \begin{tabular*}{\columnwidth}{@{\extracolsep{\fill}}lccccccc@{}}
    \toprule
    \textbf{Method} & ClipI $\uparrow$ & ClipT $\uparrow$ &  FaceSim $\uparrow$ & DINO $\uparrow$  & FID $\downarrow$ \\
    \midrule
    IP-Adapter &   86.8 & 23.3 &33.1 &92.8 & 288.5\\
    IP-Adapter$*$ &   86.9 & 23.1 &54.9(+21.8) &93.4 & 286.1\\
    InstantID & 85.8 &23.3  & 17.7  & 84.6 & 267.8\\
    InstantID$*$ &  \textbf{87.5} &23.2  &65.8(+48.1)  & 87.1 & 268.5\\
    PhotomakeV2 & \textbf{87.5}  & 23.2  &  22.5 & 87.6 & \textbf{263.1} \\
    PhotomakeV2$*$  &  \textbf{87.5}  &22.9 & 69.8(+47.3) &90.9 &  264.4\\
    PuLID &   85.7 & \textbf{24.2}  &14.2  &86.5 & 273.5\\
    PuLID$*$ &  86.1 &23.6  &68.8(+54.6)  &89.7 & 273.5\\
    \textbf{FPGA(Ours)} &  86.9  & 23.8  &  30.5 & 94.2 & 269.5 \\
    \textbf{FPGA$*$(Ours)}  &  86.9  &23.5 &\textbf{74.2}(+43.7) & \textbf{95.5} & 270.5  \\
    \bottomrule
 \end{tabular*}
 \caption{Single-ID full-body comparison, $*$ means using DIIR.}
\label{tab:Quantitative_fullbody}
\end{table}

\begin{table}[!t]
\centering\small
  \setlength{\tabcolsep}{0pt} 
  \begin{tabular*}{\columnwidth}{@{\extracolsep{\fill}}lccccccc@{}}
    \toprule
    \textbf{Method} & ClipI $\uparrow$ & ClipT $\uparrow$ &  FaceSim $\uparrow$ & DINO $\uparrow$  & FID $\downarrow$ \\
    \midrule
    IP-Adapter & \textbf{85.1} &\textbf{26.8}  &25.4  &89.7 & 140.4\\
    InstantID & 84.9 &25.9  & 30.6  &89.5 & 138.4\\
   \textbf{FPGA(Ours)}  &  84.5  &26.5 &50.1 &\textbf{92.2} &118.7   \\
    \textbf{FPGA$*$(Ours)}  &  84.6  &26.5 &\textbf{59.1}(+9.0) &\textbf{92.2} &\textbf{118.2}   \\
    \bottomrule
  \end{tabular*}
 \caption{Multi-ID full-body comparison, $*$ means using DIIR.}
\label{tab:Quantitative_MULTIid}
\end{table}

\begin{table}[!t]
\centering\small
  \setlength{\tabcolsep}{0pt} 
  \begin{tabular*}{\columnwidth}{@{\extracolsep{\fill}}lcccc@{}}
    \toprule
    \textbf{Method}  &FPGA &CodeFormer &DIIR &DIIR+CodeFormer \\
    \midrule
    FaceSim $\uparrow$ & 30.5 & 28.0  & \textbf{74.2}  & 66.7 \\
    FID $\downarrow$   & 173.5 &132.1  &153.1  & \textbf{126.9}  \\
    \bottomrule
  \end{tabular*}
 \caption{Single-ID full-body face restoration performance compared with CodeFormer. FID score is calculated only for the face region.}
\label{tab:vscodeformer}
\end{table}

\subsubsection{Evaluation Dataset}For generating single-ID half-body portraits, we selected Unsplash-50~\cite{gal2024lcm}, which consists of 50 portrait images uploaded to the Unsplash website between February and March 2024. To evaluate the capability of different methods in generating full-body images, we collect 30 full-body photographs from the Internet (with the face resolution less than 5\% of the total image), and use llava1.5~\cite{liu2023improvedllava} to generate descriptions for them. Finally, we chose 40 athlete group photos from the 2024 Paris Olympics, featuring over 100 unique IDs (including various genders, nationalities, and ages), and utilized llava1.5 to generate descriptions for these images. These images are used to test the model’s ability to generate group photo.

\subsubsection{Evaluation metrics} For quantitative comparison, we use the DINO and CLIP-I metrics to measure image fidelity, the CLIP-T\cite{DBLP:journals/corr/abs-2103-00020} metric to assess prompt fidelity, and the FID metric to evaluate the quality of the generated images. Finally, a facial recognition model~\cite{deng2020subcenter}~\cite{deng2018arcface} is employed to measure facial similarity (FaceSim). When generating multi-ID images, it is crucial to place the specified ID at the correct position. Therefore, the generated face must have an Intersection over Union (IoU) with the designated area greater than 0.5, otherwise, the similarity score is set to 0.

\subsection{Comparison Experiments}
\subsubsection{Single-ID Generation}For a fair comparison, we benchmarked against UNet-CNN based sota open-source methods, including InstantID, PuLID-SDXL, and PhotoMakerV2. We adhered to their respective official configurations by incorporating default prompts to enhance generation quality and utilized the Openpose ControlNet model to constrain poses during the generation process. On the unsplash test set, as listed in table.~\ref{tab:Quantitative_halfbody}, our method achieves the best performance across all metrics, with a significant lead in the FaceSim metric. It demonstrates that our MMF effectively integrates identity constraints with various spatial constraints in joint training, which enhances both the quality of generation and the preservation of IDs.
As shown in Fig.~\ref{fig:full_body_gen}, all the other methods failed to generate full-body portraits, as the generated faces are broken due to the limit facial resolution. As shown in Table.~\ref{tab:Quantitative_fullbody}, the similarity of the generated faces drops to below 30\%. After restoring these faces with DIIR, both the facial similarity and the DINO metrics were significantly improved, with only a minor impact on other metrics. This proves that our method can effectively repair the broken faces generated by different models without damaging the background or color information.

\subsubsection{Multi-ID Generation} The difficulty in generating multi-ID images lies in placing specific IDs in specific positions to avoid ID confusion. We configured Ip-Adapter and InstantID with proposed Mask Guided Multi-ID Cross Attention to achieve multi-ID generation, visualization results are shown in Fig.~\ref{fig:prove_CFT}. \textbf{FPGA*} indicates that we use DIIR to repair faces with resolution less than 100$\times$100. The final results in Table.~\ref{tab:Quantitative_MULTIid} show that our method leads in most metrics. Notable is the third row in Table.~\ref{tab:Quantitative_MULTIid}, where even without using DIIR, our facial similarity and generation quality still surpass other methods, demonstrating the effectiveness of our FPGA. The Clone Face Tuning strategy effectively enhances the model's capability to generate multi-ID images.

\subsection{Ablation Study}
\subsubsection{Clone Face Tuning} 
\begin{figure}[t]
\begin{center}
\includegraphics[width=\linewidth]{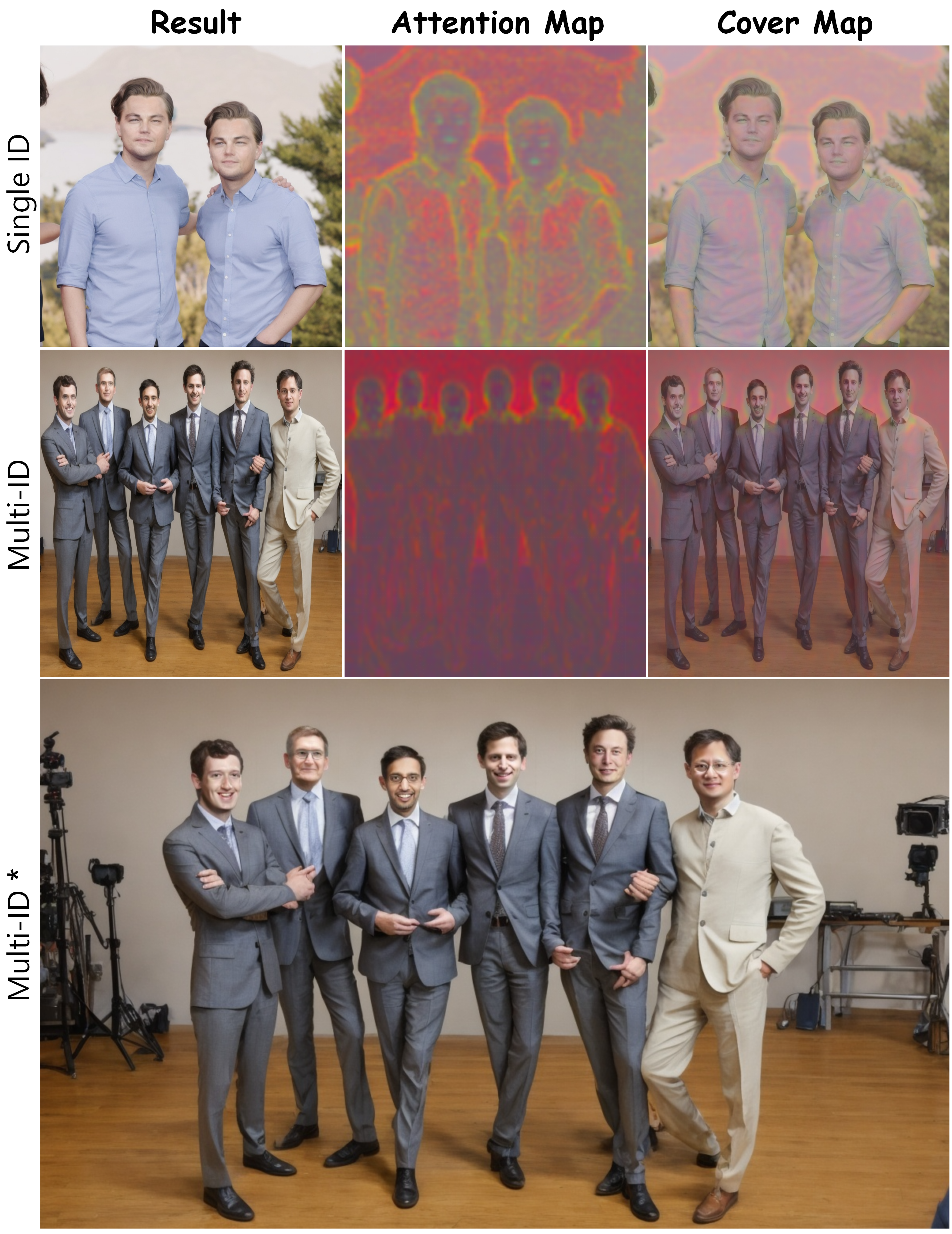}
\end{center}
   \caption{Ablation study of Clone Face Tuning, visualization of cross-attention maps and multi-ID full-body generation result. $*$ means using DIIR, more results are in supplementary materials.}
   \label{fig:multiID}
\end{figure}

\begin{figure}[t]
\begin{center}
\includegraphics[width=\linewidth]{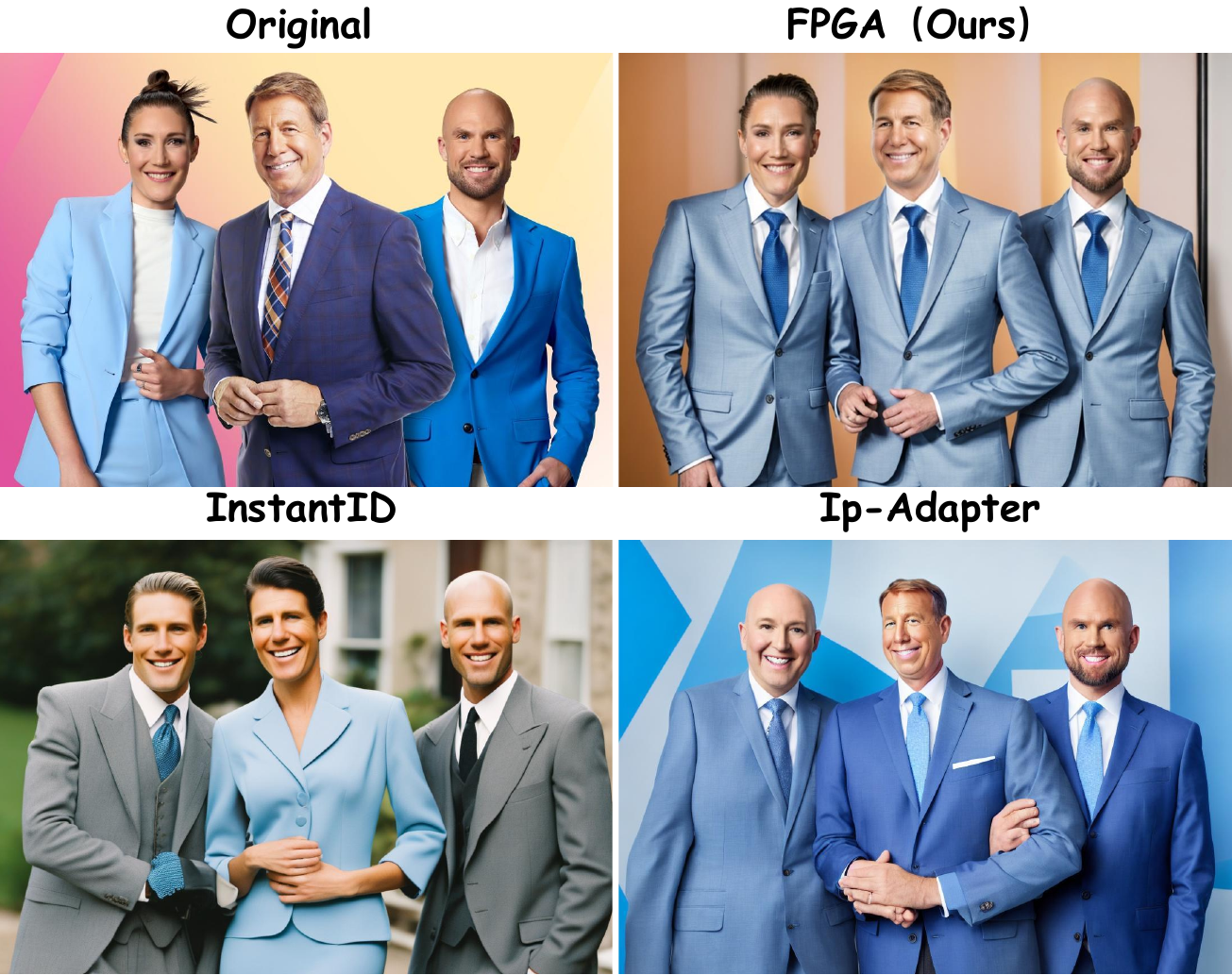}
\end{center}
   \caption{Three methods all use Mask Guided Multi-ID Cross Attention for Multi-ID generation, but only our FPGA can generate specific ID at specific region which applied Clone Face Tuning training technique.}
   \label{fig:prove_CFT}
\end{figure}

As indicated in Table.~\ref{tab:Quantitative_MULTIid}, FPGA demonstrates a strong capability in multi-ID generation. Moreover, by visualizing the Attention Map of the first layer of ControlNet, as shown in the first row of Fig.~\ref{fig:multiID}, it can be seen that a single ID can be simultaneously activated in two given facial areas, which is significant for specific ID generated at a specific position, owing to the Clone Face Tuning training technique. The activation of each ID at each given facial areas shown in second row of Fig.~\ref{fig:multiID} is significant for multi-ID generation, as shown in Fig.~\ref{fig:prove_CFT}, Instant-ID and IP-Adapter can not generate specific ID at specific region even using Mask Guided Multi-ID Cross Attention, due to not each ID can be activated at each region during inference procedure.

\begin{figure}[!t]
\begin{center}
\includegraphics[width=\linewidth]{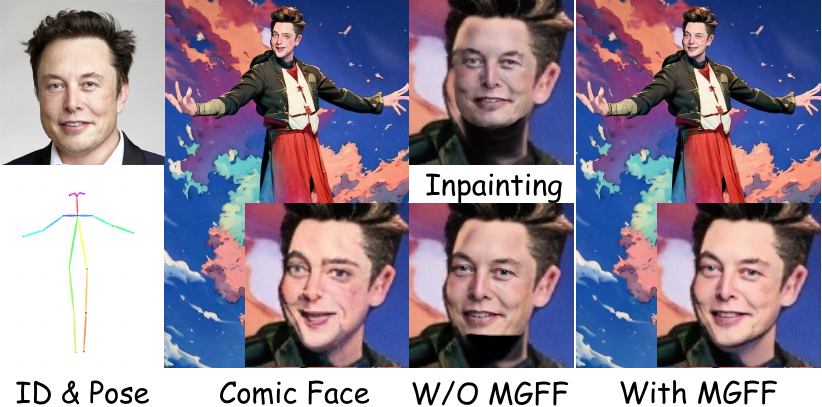}
\end{center}
   \caption{Ablation study of DIIR. Inpainting and W/O MGFF methods all fail to restore the broken face. More results are presented in supplementary materials.}
   \label{fig: Ablation study of DIIR.}
\end{figure}

\subsubsection{DDIM Inversion based ID Restoration} 
The quantitative results presented in Table.~\ref{tab:Quantitative_fullbody} have demonstrated the effectiveness of DIIR in repairing broken faces. We will further prove the effectiveness of DIIR. As shown in Fig.~\ref{fig: Ablation study of DIIR.}, we generate a full-body comic-style image of Musk with a broken face. Directly using image-to-image inpainting based on the input ID image to regenerate face region results in an asymmetrical face pose. Our Mask Guided Face Fusion can effectively extract the low-level information from the broken face to guide face restoration, and generate a face with the correct pose and harmonious style. In Table~\ref{tab:vscodeformer}, we compare our DIIR with the sota face restoration method CodeFormer~\cite{codeformer}. Although CodeFormer can enhance the quality of low-resolution face, it is unable to recover damaged identity information. However, our DIIR is capable of simultaneously repairing identity information and improving face quality. By employing CodeFormer and DIIR simultaneously, we can achieve the goal of recovering identity information while further enhancing face quality.

\subsection{Acceleration Evaluation} 
We further conduct Identity-Preserving generation quantitative experiments to verify the portability of our RepControlNet. We evaluated the identity-preserving generation capabilities of various models on the Unsplash-50~\cite{gal2024lcm} dataset by comprehensively assessing their performance through metrics such as CLIP scores, facial similarity(FaceSim), and FID. We benchmarked against sota open-source solutions, including InstantID~\cite{wang2024instantid}, PuLID ~\cite{guo2024pulid}, and PhotoMakerV2~\cite{li2023photomaker}. To ensure a fair comparison, we adhered to their respective official configurations by incorporating default prompts to enhance generation quality and utilized the Openpose ControlNet model to constrain poses during the generation process. On the unsplash test set, as listed in Table.~\ref{tab:Quantitative_halfbody}, our RepControlNet achieved the best performance across all metrics, with a significant lead in the FaceSim metric. These results fully validate the portability of our proposed RepControlNet from objective metrics.
\begin{table}[!t]\centering
    \begin{tabular}{c|ccc}
        \hline
        Method & FLOPs & MACs & Param. \\ 
        \hline
        SD1.5  & 0.68 T & 0.34 T  & 1066 M \\
        FPGA(ControlNet+SD1.5) & 0.91 T  & 0.46 T & 1427 M \\
        \textbf{FPGA(RepControlNet)} & \textbf{0.69 T} & \textbf{0.35 T} & \textbf{1067 M} \\
        \hline
    \end{tabular}
    \caption{Comparison of FLOPs, MACs and parameters between before and after reparameterization.}
    \label{table:compute}
\end{table}


\begin{table}[!t]
\centering
  \setlength{\tabcolsep}{0pt} 
  \begin{tabular*}{\columnwidth}{@{\extracolsep{\fill}}c|cccc@{}}
    \hline
    \textbf{Method}   & CLIP-score$\uparrow$ &  FaceSim$\uparrow$   & FID$\downarrow$ \\
    \hline
    IP-Adapter  & 0.20 &41.1 & 213.8\\
    InstantID   & 0.19  &57.7 &220.3\\
    PhotomakeV2  & \textbf{0.21}  &  50.5 &  216.1 \\
    PuLID& 0.19 & 47.0  & 229.9\\
    RepControlNet &  \textbf{0.21}  & \textbf{60.7}  &\textbf{195.8} \\
    \hline
  \end{tabular*}
 \caption{Skeleton guided portrait fidelity generation quantitative comparison. The best average performance is in bold. ↑ indicates higher metric value and represents better performance and vice versa.}
\label{tab:Quantitative_halfbody}
\end{table}
\section{Analysis and Conclusions}

Our FPGA achieves a more flexible portrait photo generation ability, which can satisfy the multi-ID position-specified portrait generation task.
We propose the Multi-Mode Fusion training strategy (MMF) to improve the generalization ability of multi-ID generation. The Clone Face Tuning and Mask Guided Multi-ID Cross Attention are introduced to enhance the control of specific ID generated at specific region during the inference stage.
DIIR is proposed to restore and enhance the face details when the generated face resolution is too small. The DIIR is plug-and-play and can be incorporated with any diffusion-based portrait generation methods to further enhance their performance.
Extensive experiments demonstrate that FPGA not only achieves superior performance in objective and subjective metrics, but also can be widely extended to other portrait generation methods. And after our RepControlNet engineering acceleration, it can complete an inference process within 2.5 seconds on a L20 graphics card.


\bibliographystyle{IEEEtran}

\appendix
\begin{figure*}[!t]
\begin{center}
\includegraphics[width=0.88\textwidth]{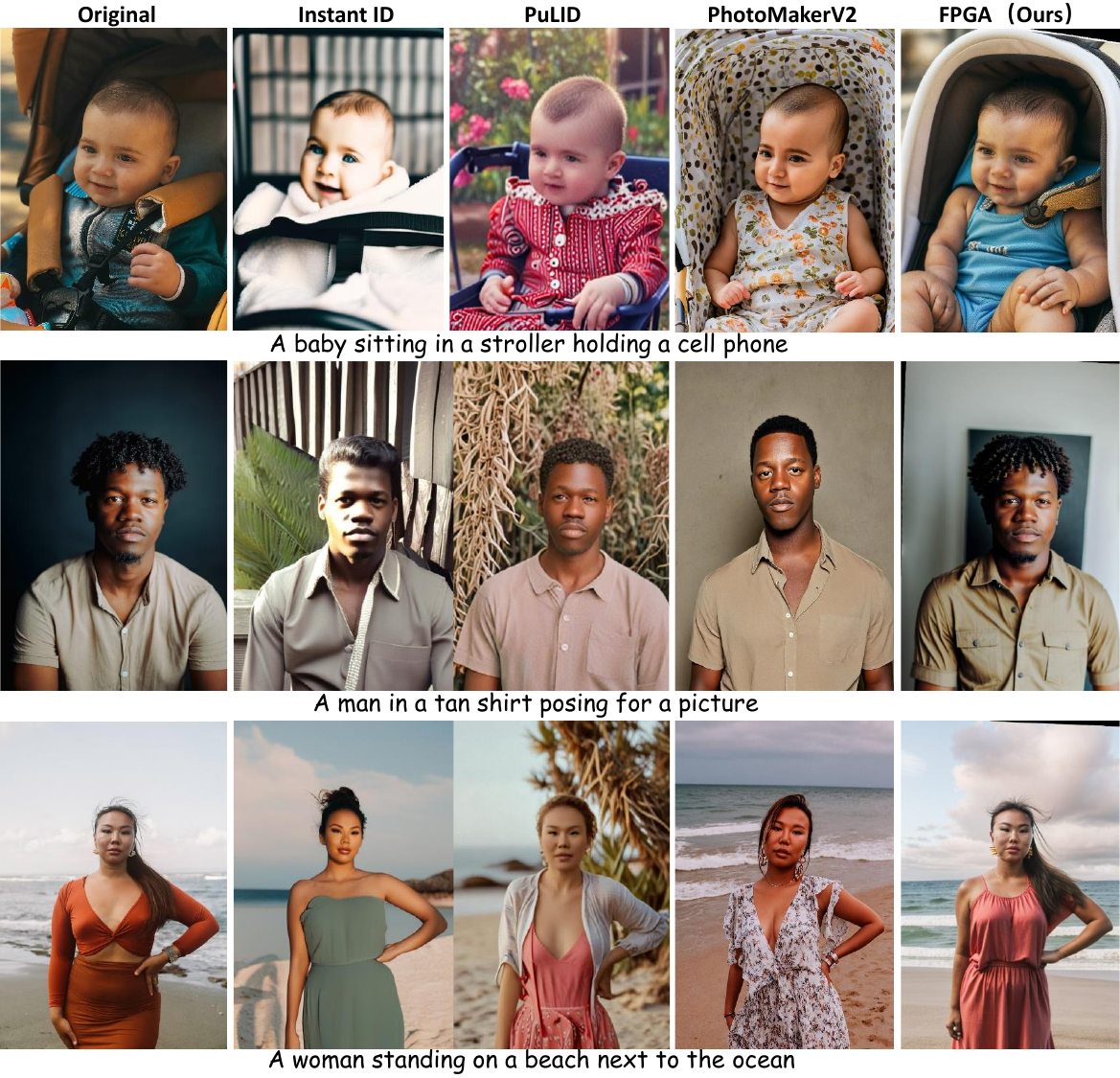}
\end{center}
   \caption{Visualization of Single-ID half-body generation results.}
   \label{fig:sing_ID1}
\end{figure*}

\begin{figure*}[t]
\begin{center}
\includegraphics[width=0.88\textwidth]{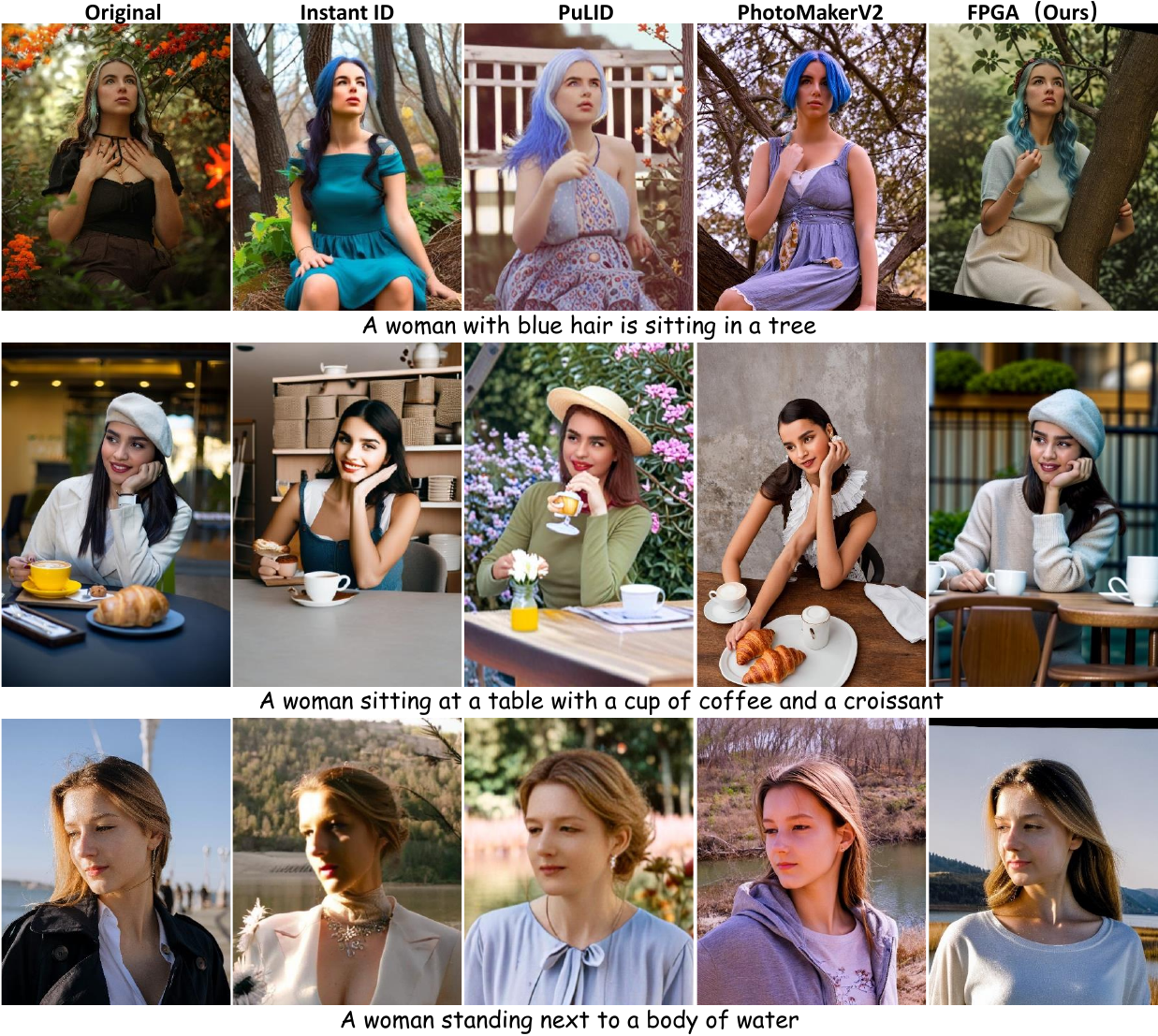}
\end{center}
   \caption{Visualization of Single-ID half-body generation results.}
   \label{fig:sing_ID2}
\end{figure*}

\begin{figure*}[t]
\begin{center}
\includegraphics[width=0.88\textwidth]{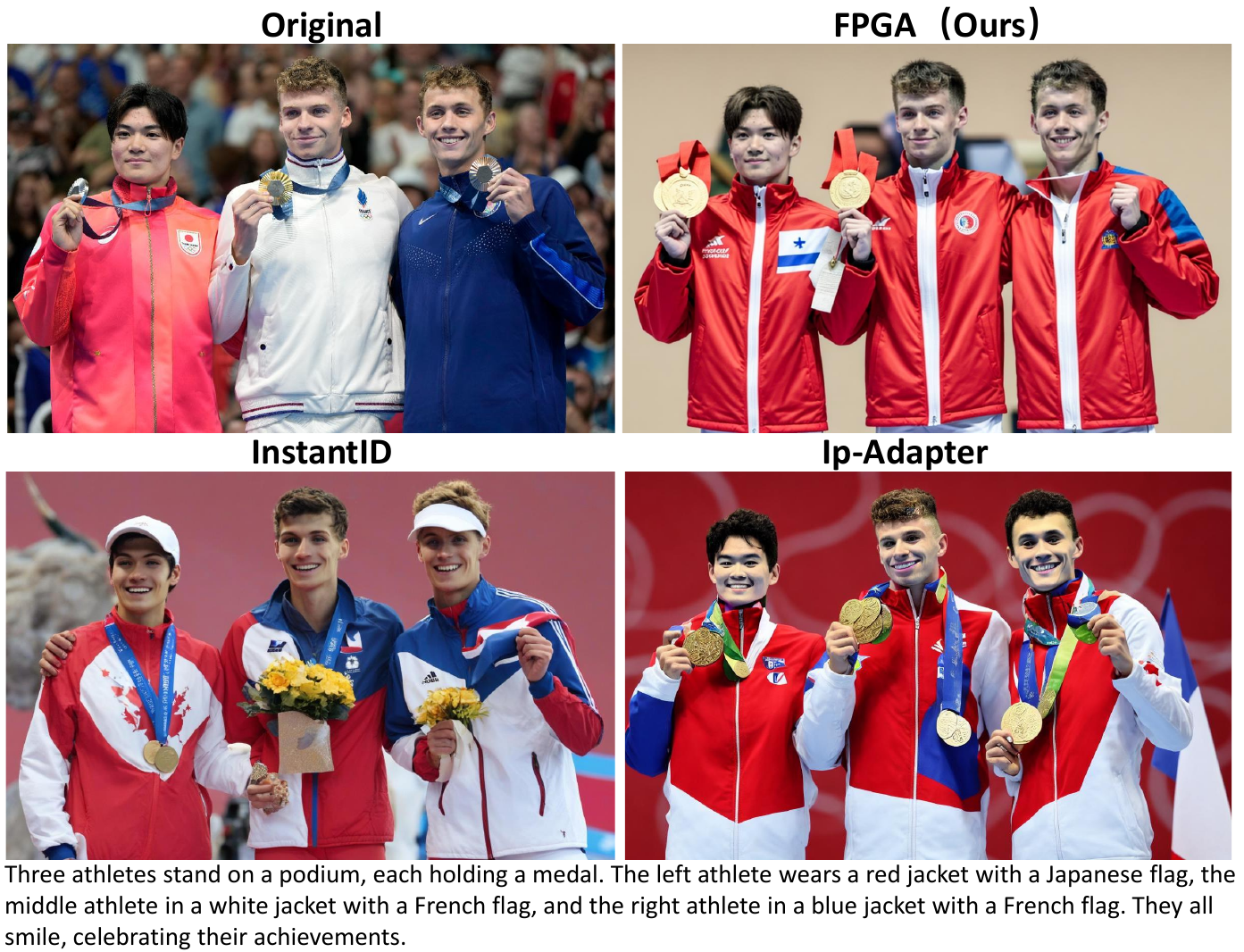}
\end{center}
   \caption{Visualization of Multi-ID half-body generation results.}
   \label{fig:MULTI_ID_HALF_BODY1}
\end{figure*}

\begin{figure*}[t]
\begin{center}
\includegraphics[width=0.88\textwidth]{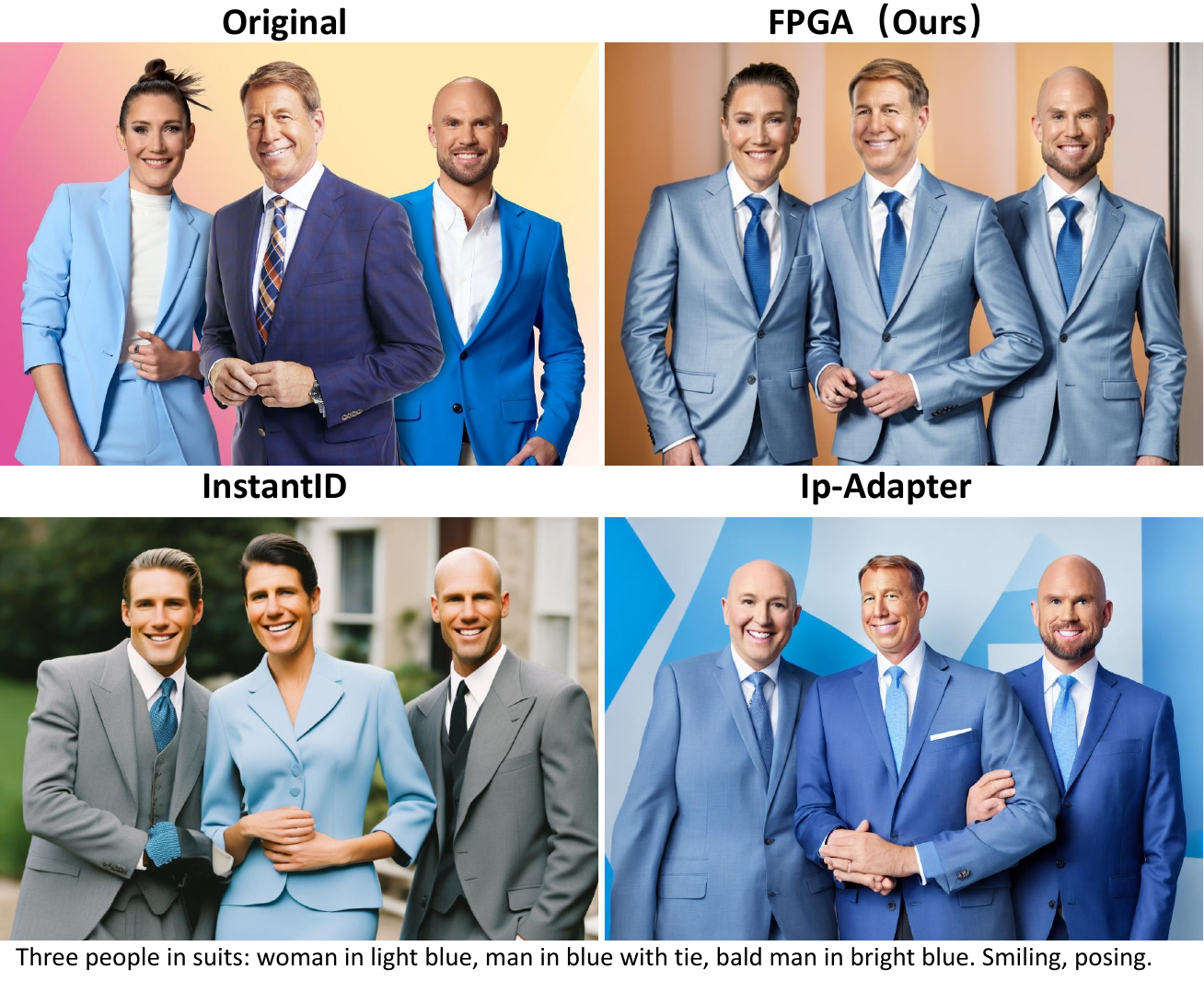}
\end{center}
   \caption{Visualization of Multi-ID half-body generation results}
   \label{fig:MULTI_ID_HALF_BODY2}
\end{figure*}

\begin{figure*}[t]
\begin{center}
\includegraphics[width=0.88\textwidth]{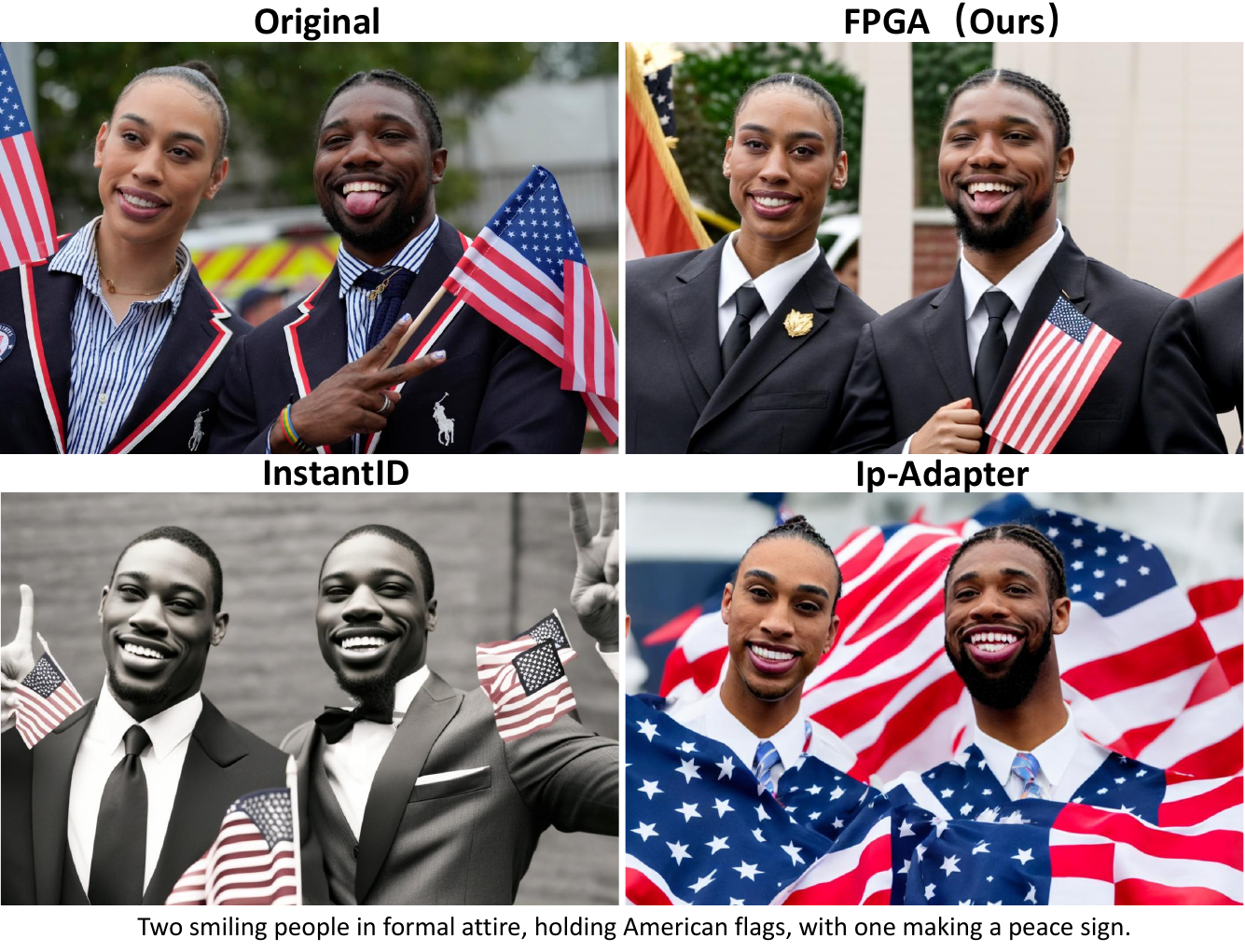}
\end{center}
   \caption{Visualization of Multi-ID half-body generation results.}
   \label{fig:MULTI_ID_HALF_BODY3}
\end{figure*}

\begin{figure*}[t]
\begin{center}
\includegraphics[width=0.88\textwidth]{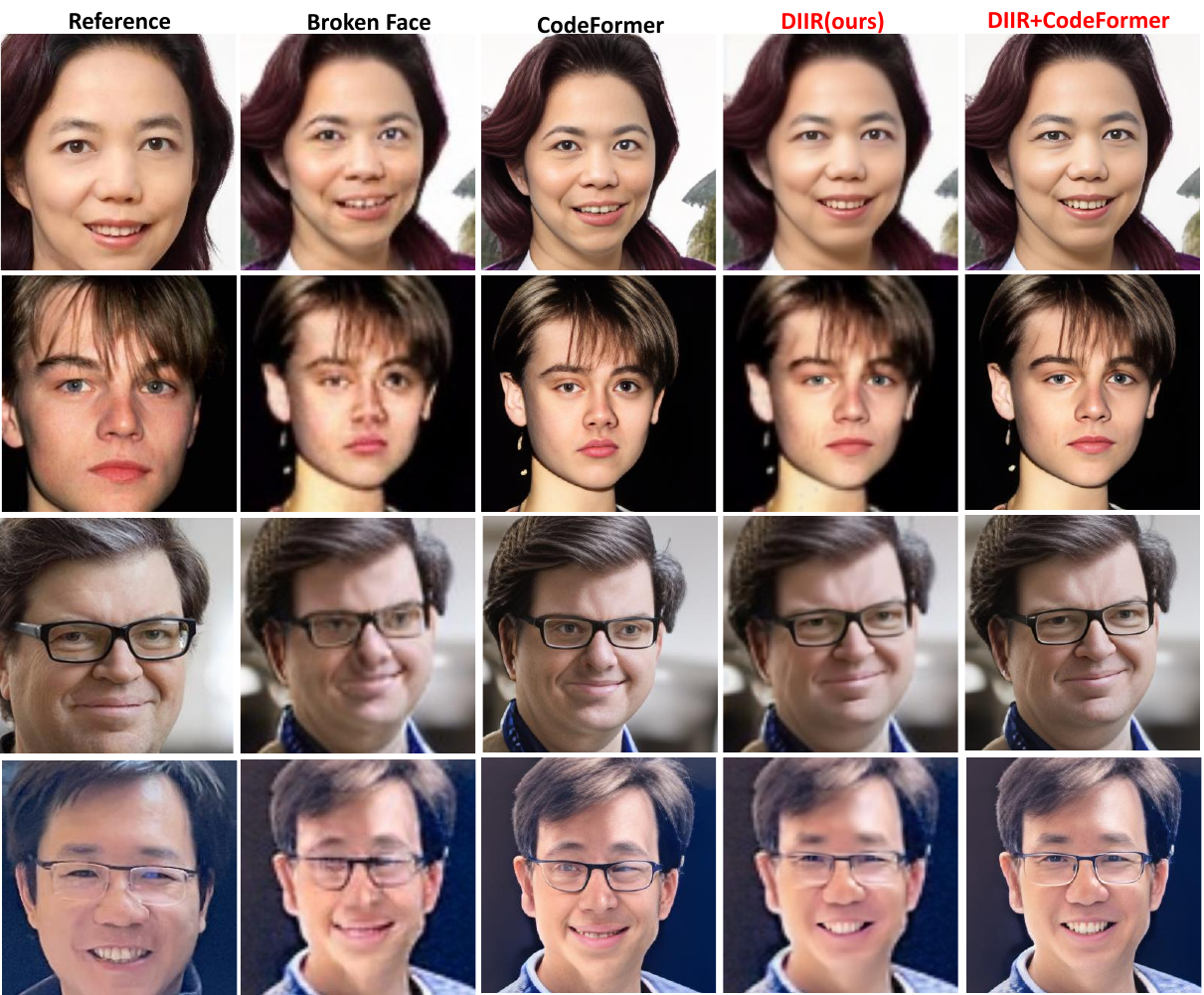}
\end{center}
   \caption{DIIR can be easily combined with CoderFomer, to further enhance the quality of generated faces.}
   \label{fig:DIIR_CodeFomer_face}
\end{figure*}

\begin{figure*}[t]
\begin{center}
\includegraphics[width=0.88\textwidth]{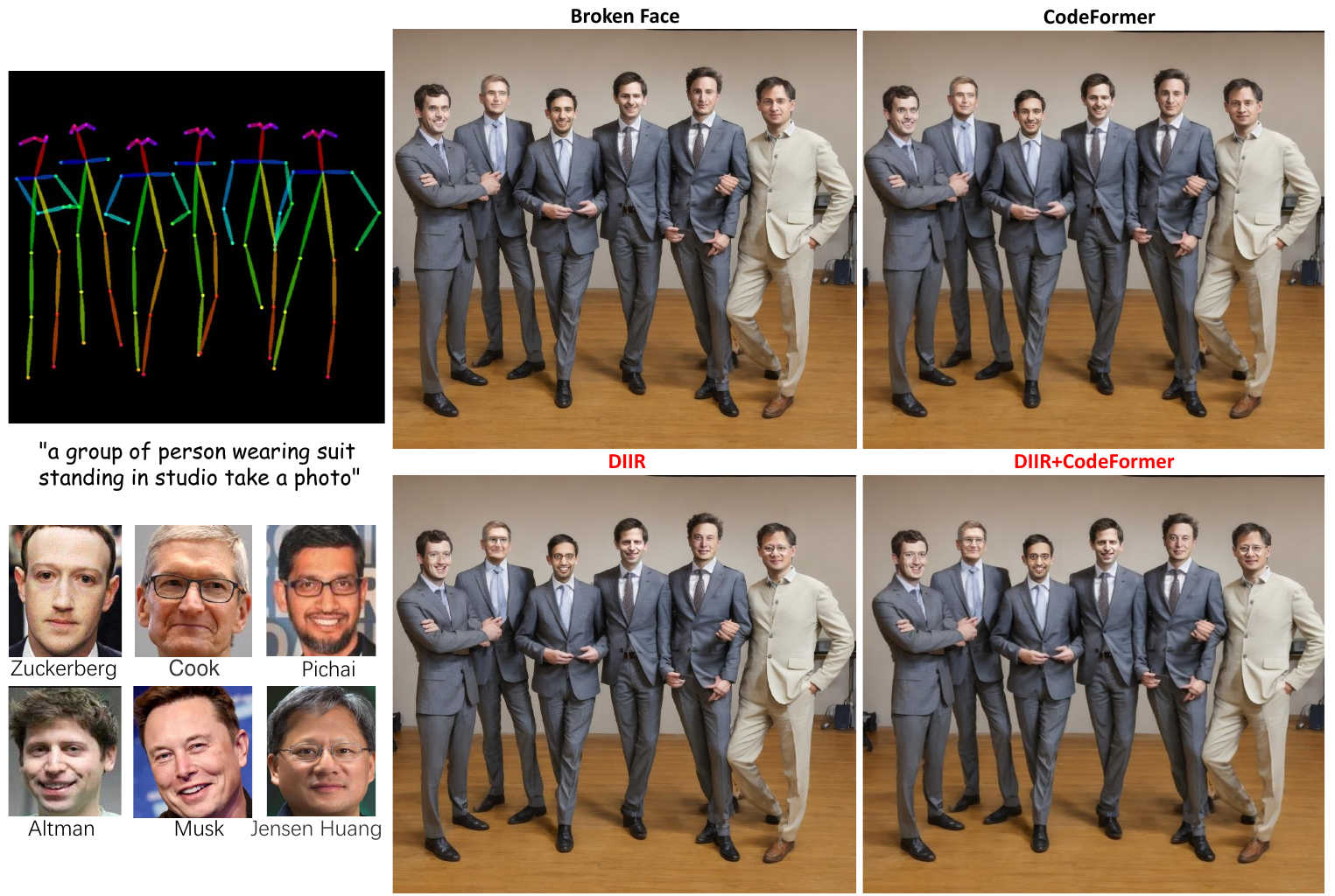}
\end{center}
   \caption{Visualization of Multi-ID full-body generation results(most challenging task). DIIR combines CoderFomer can greatly boost the ability of this issue. }
   \label{fig:MULTI_ID_FULL_BODY1}
\end{figure*}

\begin{figure*}[t]
\begin{center}
\includegraphics[width=0.88\textwidth]{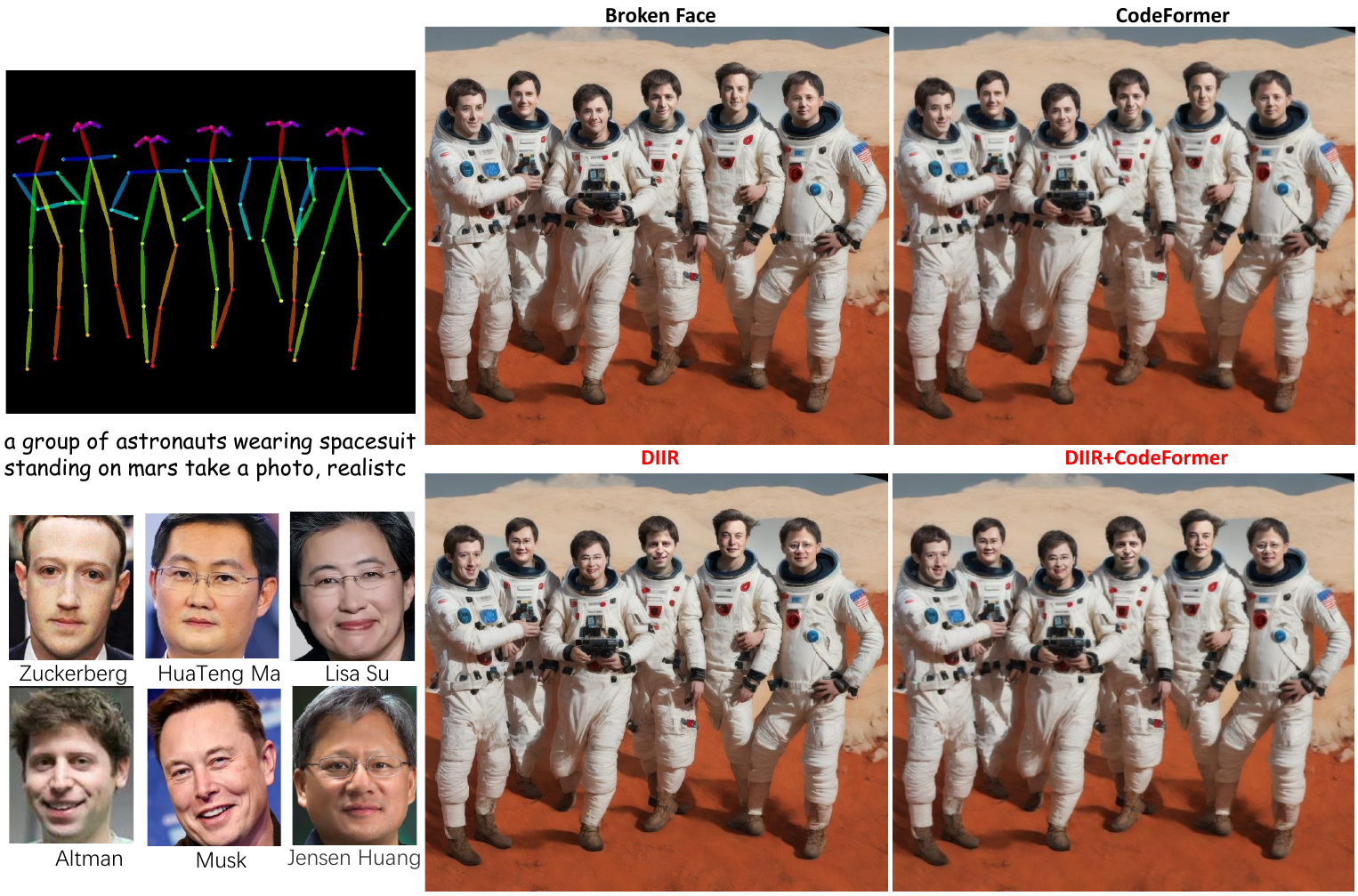}
\end{center}
   \caption{Visualization of Multi-ID full-body generation results(most challenging task). DIIR combines CoderFomer can greatly boost the ability of this issue.}
   \label{fig:MULTI_ID_FULL_BODY2}
\end{figure*}

\begin{figure*}[t]
\begin{center}
\includegraphics[width=0.88\textwidth]{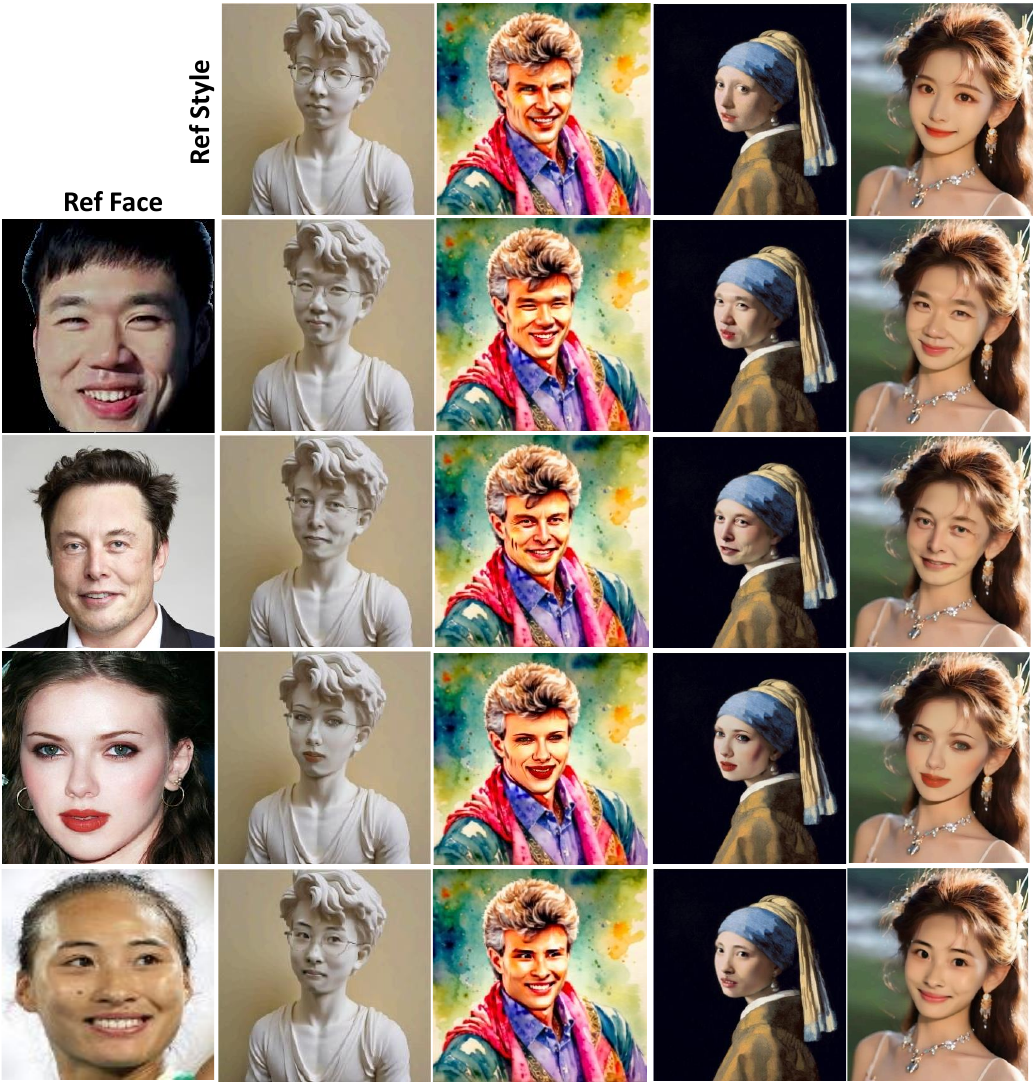}
\end{center}
   \caption{DIIR can combined with stylization methods, and have the ability of swapping stylized face while keeping the expression.}
   \label{fig:DIIR_Style}
\end{figure*}





\end{document}